\documentclass[11pt]{article}
\usepackage{times}
\usepackage[letterpaper, width=6.5in, height=9.0in]{geometry}
\usepackage{comment}
\usepackage{microtype}
\usepackage{graphicx}
\usepackage{subcaption}
\usepackage{booktabs} % for professional tables
\usepackage{bm}
\usepackage{multirow}

\usepackage{hyperref}

\usepackage{xcolor}
\newcommand\gk[1]{{\textcolor{black}{#1}}}
\newcommand\na[1]{{\color{black}#1}}

\usepackage{array,collcell}
\newcommand\elbowK{0}
\newcommand\elbowcell[1]{\ifnum#1=\elbowK\relax\fbox{\makebox[1.2em][c]{\strut#1}}\else#1\fi}
\newcolumntype{E}{>{\collectcell\elbowcell}c<{\endcollectcell}}

\newcommand{\MNTenAETrainLoss}{0.0770}
\newcommand{\MNTenAETestLoss}{0.0775}
\newcommand{\MNTenAEBestEpoch}{98}
\newcommand{\MNTenAEEpochs}{100}
\newcommand{\MNTenBestK}{20}
\newcommand{\MNTenBestNMI}{0.690}
\newcommand{\MNTenBestARI}{0.472}
\newcommand{\MNTenBestACC}{0.856}
\newcommand{\MNTenKneeK}{8}
\newcommand{\MNTenBICK}{20}

\newcommand{\MNFortyAETrainLoss}{0.0699}
\newcommand{\MNFortyAETestLoss}{0.0697}
\newcommand{\MNFortyAEBestEpoch}{100}
\newcommand{\MNFortyAEEpochs}{100}
\newcommand{\MNFortyBestK}{80}
\newcommand{\MNFortyBestNMI}{0.695}
\newcommand{\MNFortyBestARI}{0.414}
\newcommand{\MNFortyBestACC}{0.753}
\newcommand{\MNFortyKneeK}{35}
\newcommand{\MNFortyBICK}{80}
\newcommand{\MNFortyPenalizedK}{30}

\usepackage{enumitem}

\usepackage{amsmath}
\usepackage{amssymb}
\usepackage{mathtools}
\usepackage{amsthm}

\usepackage[capitalize,noabbrev]{cleveref}

\theoremstyle{plain}

\theoremstyle{definition}

\theoremstyle{remark}

\usepackage[textsize=tiny]{todonotes}

\newcommand{\squishlist}{
   \begin{list}{$\bullet$}
    { \setlength{\itemsep}{0pt}      \setlength{\parsep}{0pt}
      \setlength{\topsep}{3pt}       \setlength{\partopsep}{0pt}
      \setlength{\listparindent}{-2pt}
      \setlength{\itemindent}{-5pt}
      \setlength{\leftmargin}{1em} \setlength{\labelwidth}{0em}
      \setlength{\labelsep}{0.5em} } }

\newcommand{\squishend}{
    \end{list}  }

\title{Autoencoder Based Deep Clustering of Point Clouds}

\author{Nader Alfares, Guangmingmei Yang, David J. Miller, George Kesidis\\
School of EECS, Penn State, University Park, PA, USA\\
\{nna5040,gzy5102,djm25,gik2\}@psu.edu}

\begin{document}
\maketitle

\begin{abstract}
Motivated 
by a problem in
Generative Specification-Producing AIs (SPAIs),
we focus on a problem of deep clustering of three-dimensional
shapes specified by point clouds. After reviewing clustering and
deep clustering, we study the fully unsupervised pipeline in which a
FoldingNet autoencoder is first trained for reconstruction and then
$K$-means++ is then applied to the resulting codewords. We report clustering accuracy (ACC), normalized mutual information (NMI) and adjusted Rand index (ARI)  on ModelNet10 and ModelNet40 datasets, as a function of the number of clusters $K$. We also provide t-SNE visualizations of the codewords, and discuss the choice of $K$.
\end{abstract}

\section{Introduction}

We herein focus on the problem of deep clustering of three-dimensional shapes.
Our motivation is the problem of  
Generative Specification-Producing AIs (SPAIs) \cite{SPAI}.
SPAIs may have multimodal prompts 
(as transformer VLMs \cite{singh2018attentionattentionarchitecturesvisual})
with diagrams (shapes) and text conveying design requirements.
That is, a description of a new design is input and a ``formal" design specification is output.
The output specification is informed by the designs used to train (or fine-tune) a supervised-learned SPAI.
A good design specification for a given set of requirements will be informed by similar ones used to train the SPAI.
So, we are concerned with feature maps (encoders) that are useful for multimodal clustering to identify the similar designs used for training, i.e., a retrieval function  \cite{Joshi21}.

In this paper, we focus our study on deep clustering of 3D shapes, which are typically specified by point clouds or tesselaations on their surfaces. These two formats can be translated from one to the other. Moreover, three-dimensional shapes are invariant to
point-order specification \cite{PointNet}, translations,  rotations, and uniform dilations (or contractions)
\cite{Small96}. Hence, for consistency and simplicity, we focus our study on point clouds.   In particular, we study the deep clustering of point clouds using a FoldingNet autoencoder \cite{yang2018FoldingNet}. 

This paper is organized as follows.
We first give an overview of $K$-means clustering and of deep
clustering, in which a learned feature map (encoder) is followed by
clustering in the embedded feature (latent, activation) space, and
of the cluster collapse to which \textit{deep} clustering is
particularly prone. We then describe the ModelNet datasets and the
clustering metrics used, describe the autoencoder/$K$-means++
pipeline, and report experimental results on ModelNet10 and
ModelNet40.

The main \gk{contribution} is an empirical study of the fully
unsupervised autoencoder/$K$-means++ pipeline for deep clustering of
shapes: a FoldingNet autoencoder \cite{yang2018FoldingNet} is trained
for reconstruction only, and $K$-means++ is applied to its codewords. We report
results for experiments conducted using the ModelNet10 and ModelNet40
datasets \cite{ModelNet}, which pose challenges because they are
highly imbalanced, examine the effect of the number of clusters $K$ on
the clustering metrics for both datasets, and analyze qualitatively
which shape categories are merged in codeword space.

\section{Background on Clustering}

\begin{figure}
    \centering
    \includegraphics[width=0.5\columnwidth]{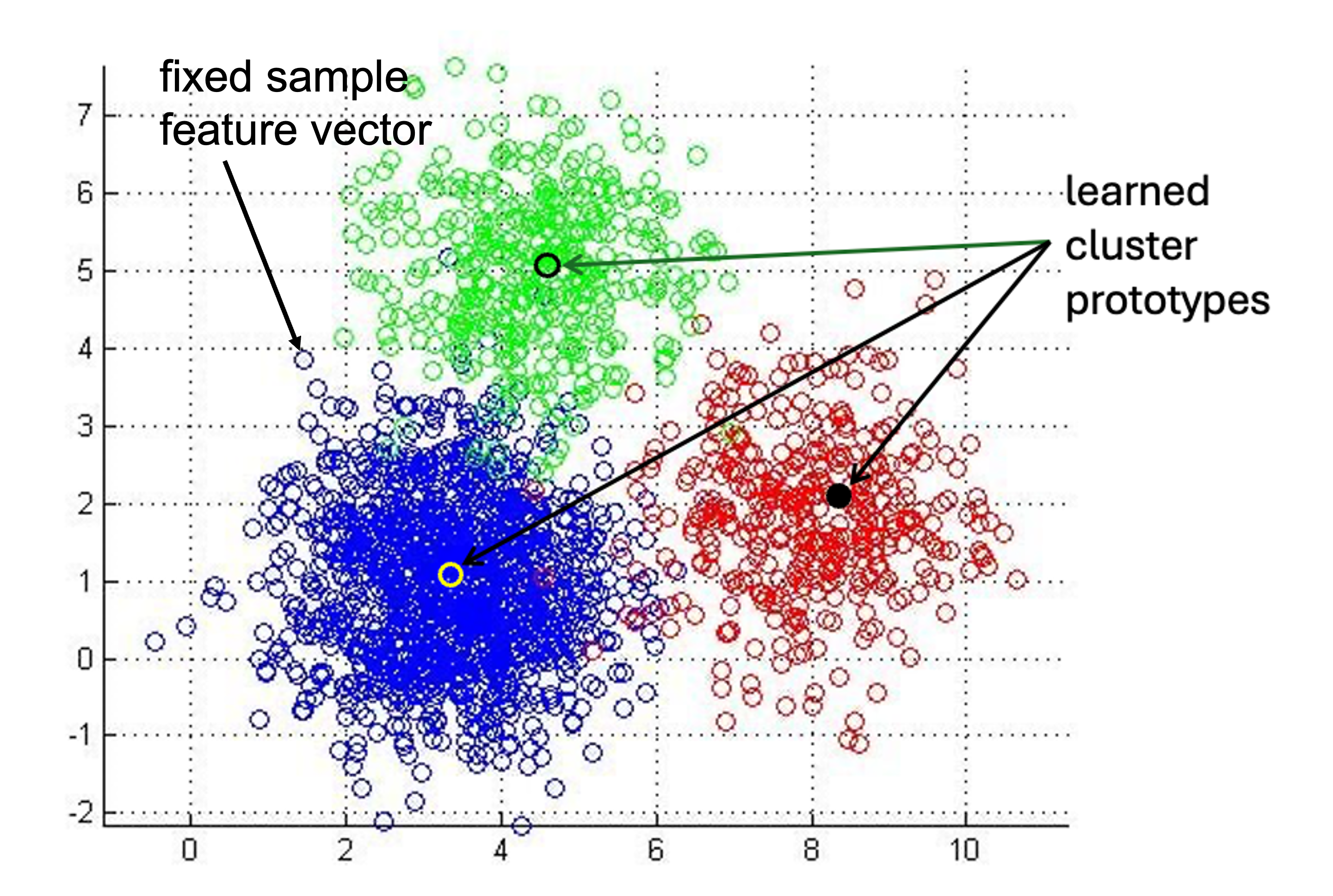}
    \caption{Illustrative example of clustering scenario with $K=3$ clusters.}
    \label{fig:clustering_illus}
\end{figure}

Given an unlabeled training set $X$ of point clouds, one can apply $K$-means clustering
(e.g., \cite{Duda}) using the Chamfer distance,
 $d_{\rm c}$, between pairs of point clouds.
Given $X$, $K$-means clustering uses an iterative expectation-maximization 
(map-reduce) to determine the $K$ cluster prototypes $\{\pi_1,...,\pi_K\}$ by minimizing
the cluster-distortion loss,
\begin{align*}
L_0(\pi) & = \frac{1}{|X|}\sum_{x\in X} d_{\rm c}(x,\pi(x)), ~~\text{where}\\
\pi(x) & =\text{arg}\min_{\pi_m} d_{\rm c} (x,\pi_m) ~~\text{and, e.g.,}\\
d_{\rm c}(x,y) & = \frac{1}{|x|} \sum_i  \min_j \| x_i-y_j\|
+ \frac{1}{|y|} \sum_j  \min_i \| y_j-x_i\|
\end{align*}
where here $x_i,y_j \in\mathbb{R}^3$ respectively are the points of (point clouds) $x,y$
and $\|\cdot\|$ is the Euclidean norm.
See Figure \ref{fig:clustering_illus} for a simple illustrative 
example with $K=3$ clusters.

Clearly, the cluster-distortion loss can be zero when $K\geq |X|$.
To optimize generalization performance (i.e., avoid overfitting),
the number of clusters $K$ can be set using BIC \cite{Schwarz78,CI}.
\na{Treating $K$-means as a spherical Gaussian mixture with shared
variance, the BIC of a partition with residual mean square error $L(\pi)$ (see equ. \eqref{eq:kmeans} below)
%$\mathrm{SSE}$ 
is
\begin{equation}\label{eq:bic_gmm}
ND\ln(L(\pi)/D)+KD\ln N,
\end{equation}
where $N=|X|$ and $D$ is the common or average dimension of each
encoded shape representation, where lower is better. A simpler additive form
adds only the complexity term}
\begin{equation}\label{eq:BIC}
\frac{1}{2}KD \log_2 |X|
\end{equation}
to the cluster-distortion loss $L_0$ or $L$.
Alternatively,
one can apply a heuristic approach such as the ``elbow" method
\na{\cite{Thorndike53}: a clustering criterion that improves
monotonically with $K$, classically the cluster-distortion loss $L_0$,
is plotted against $K$, and $K$ is set to the smallest value beyond
which the marginal improvement obtained by adding one more cluster
becomes small, i.e., to the ``elbow'' of the curve, past which the
returns of increasing $K$ diminish.}

On the other hand,
\textbf{cluster collapse} is a phenomenon where all the training samples 
are assigned to only a few (even just one) cluster. $K$-means may exhibit cluster
collapse, e.g., when the cluster prototypes are poorly initialized.

\textbf{Deep} clustering involves a front-end feature map (encoder) and 
a back-end clustering.
The motivation for applying clustering within a deep learning 
setting is that, for certain types of data, clustering applied directly to the ``raw'' input data may produce poor results.  For example, consider object clustering in image domains.  In different images, the same object may appear in different locations, at different scales, from different perspectives, and in the presence of noise and occlusion, within an image dataset.  Applying clustering directly to the raw images is not likely to assign images representing the same object (but subject to these highly variable factors) to the same cluster.  However, a deep neural network can learn embedded feature representations that are relatively invariant to these variable factors.  Thus, a clustering method applied to deep learning embedded representations of images is much more likely to assign images containing the same object to a common cluster. 

As another example, an
attempt to detect poisoned samples in the training dataset of a DNN-based classifier 
has been based on clustering of embedded representations of
the training data samples \cite{SS,BChen18,CI}.
\cite{SS,BChen18} employ a two-component clustering approach (simple 2-means clustering in the case of \cite{BChen18}), while \cite{CI} employs a Gaussian mixture model whose number of components is determined by BIC. 
As another example, \cite{KJ12} also considers mixture models and penalizes the number of clusters.

Cluster collapse is more problematic for deep clustering
when the feature map (encoder) and cluster prototypes
(in the range of the feature map) are \textit{jointly} learned: if the
encoder is trained against a cluster-distortion objective, it can
trivially reduce the distortion by mapping all inputs to (nearly)
the same codeword\footnote{A semisupervised approach can employ pairs of training examples
that must or cannot be in the same cluster, with corresponding contrastive-loss
terms in the training objective.}. 
The fully unsupervised approach studied herein
avoids this by decoupling the two problems. The encoder is first
learned as the front end of an autoencoder
\cite{yang2018FoldingNet} trained for reconstruction only, an
objective that does not involve the prototypes and that forces
the codeword to retain enough information about the input to
reconstruct it, so that distinct shapes cannot all be mapped to the
same codeword. The encoder is then fixed and $K$-means is applied
to the resulting codewords, with the prototypes initialized by
$K$-means++ seeding to guard against the collapse of $K$-means
itself.

Regarding deep clustering of shapes in particular, the encoder
should respect the invariances of three-dimensional shapes noted in
the Introduction. Point-order invariance is built into point-cloud
encoders such as PointNet \cite{PointNet} and FoldingNet
\cite{yang2018FoldingNet} through a symmetric (max) pooling over
points, while invariance to translation, rotation and uniform
dilation (or contraction) is obtained by normalizing the shapes
and, for the pre-aligned ModelNet variants used herein
(Section~\ref{sec:datasets}), by the canonical orientation of the
data\footnote{Alternatively, the training dataset can be augmented 
by rotations, scalings and translations its samples.}.

\section{ModelNet Datasets for Clustering}
\label{sec:datasets}

\begin{figure}
    \centering
    \includegraphics[width=\columnwidth]{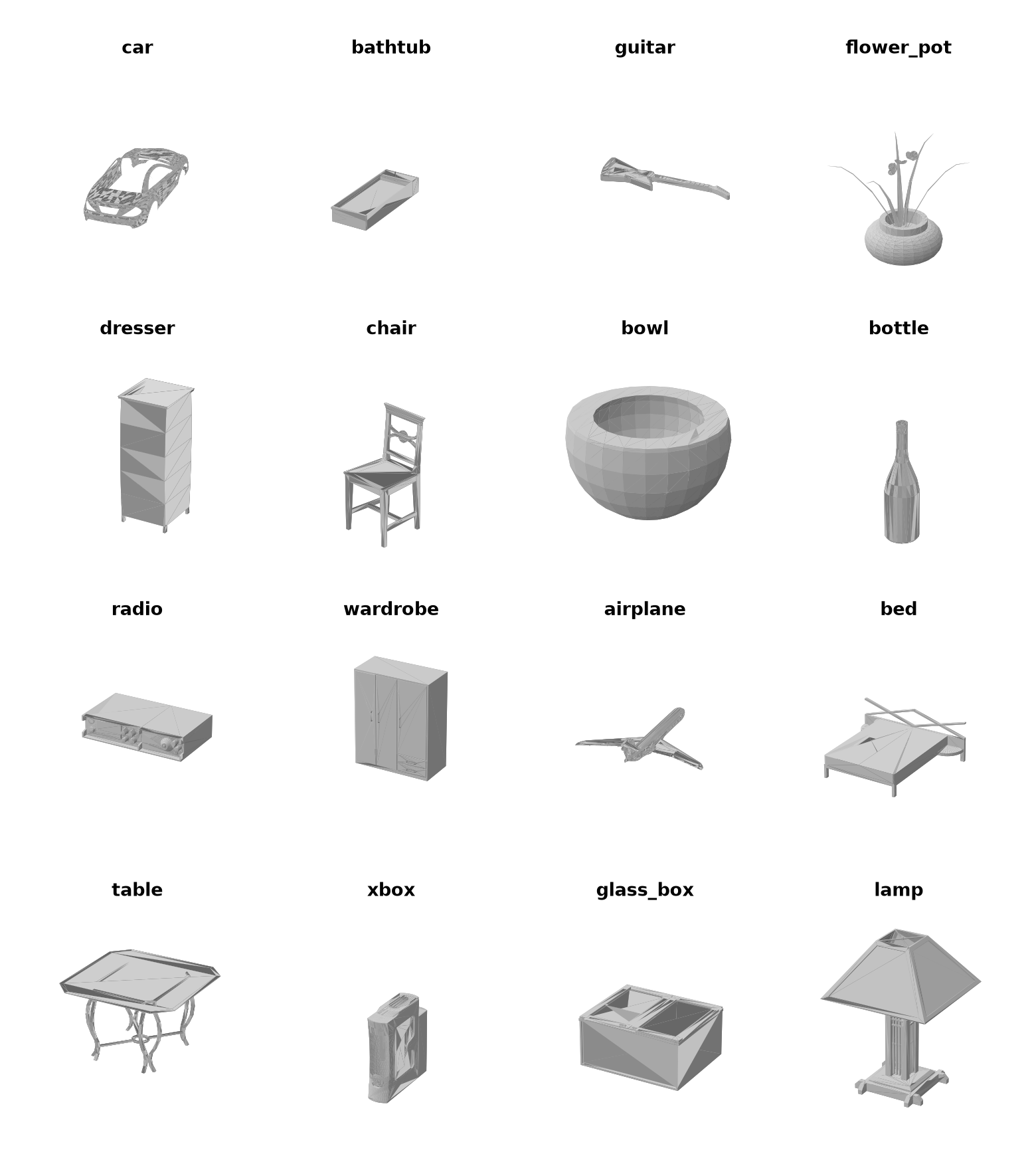}
    \caption{Some example ModelNet shapes and their labeled category.}
    \label{fig:ModelNet}
\end{figure}

We use the two standard ModelNet benchmarks \cite{ModelNet}.
ModelNet40 (MN40) consists of 12,311 shapes from 40 categories
(classes), officially split into 9,843 (80\%) for training and 2,468
(20\%) for testing; we use the pre-aligned (``auto-aligned'') variant
in which all shapes of a class share a canonical orientation.
ModelNet10 (MN10) is a subset of ModelNet40 containing 4,899 shapes
from 10 categories (bathtub, bed, chair, desk, dresser, monitor, night
stand, sofa, table, toilet). In both datasets each shape is a 3D
triangle (surface) mesh in the
%Object File Format
.OFF format \gk{(see Figure \ref{fig:ModelNet}),} which we normalize
by inscribing in a unit sphere and convert to a point cloud by
uniformly sampling points from the surface. Both datasets are highly
imbalanced: in MN10 the largest class (chair) has 989 shapes and the
smallest (bathtub) 156, and in MN40 the class sizes range from 989
(chair) down to 84 (bowl) and 99 (cup), so the imbalance is more
severe and the classes more numerous and fine-grained (e.g., desk,
table, night stand and dresser, or flower pot, plant and vase).

For unsupervised clustering, we treat both datasets identically: we
combine ModelNet's official test and training sets and
re-partition the union at random into an 80\% training split
and a 20\% held-out split (fixed random seed), which gives 3,920 /
979 shapes for MN10 and 9,849 / 2,462 shapes for MN40. The autoencoder is
trained on the 80\% split and its reconstruction loss is monitored on
the 20\% held-out split to detect overfitting of the encoder.
Since the class labels are never used for training, $K$-means++ and
the clustering metrics below are then computed on the codewords of
all shapes (4,899 for MN10 and 12,311 for MN40).
%and then randomly partitioned into, e.g., 5 equally sized folds, where 4 folds them are used to train while the 5th for testing/evaluation.

%To assess clustering accuracy after training,
%training-set class plurality is ascertained for each cluster 
%(i.e., using the class labels which were \textbf{not} used for training).
%That is, if the plurality of training samples clustered to
%prototype $\pi$ belong to class $\kappa(\pi)$, 
%\gk{and} $c(x)$ is the ground-truth
%class of point cloud $x$, then the estimated error rate is
%\begin{align}\label{eq:ACC_0}
%\frac{1}{|X|}\sum_{x \in X}
%{\bf 1}\{c(x)\not = \kappa(\pi(x))\}
%\end{align}
%where $X$ is the full set of (unlabeled) shapes that were clustered.
%\gk{Performance metrics are discussed further below.}

%Rather than adopting the recommended train/test partitions, we construct our own 80/20 split of each dataset. This choice ensures a consistent and representative distribution of object categories across both the training and testing sets, which the original splits do not guarantee.

We evaluate our approach using the following metrics:
\squishlist
    \item \textbf{Accuracy (ACC)}: Clustering accuracy measures the percentage of samples correctly assigned to clusters using a plurality matching strategy. For each predicted cluster $j$, we find the most common (plurality) true label within that cluster, then count how many samples in the cluster have that label.
    %\gk{To clarify \eqref{eq:ACC_0}}, 
    Let $\mathcal{C}_j$ % = \{i : c_i = j\}$ 
    be the set of samples \gk{indexed $i$} assigned to cluster $j$.
    %and let $y_{\mathcal{C}_j}$ be the ground-truth labels of samples in cluster $j$. 
    The plurality label for cluster $j$ is:
    \begin{equation}
    y_j^* = \arg\max_{y \in \mathcal{Y}} |\{i \in \mathcal{C}_j : y_i = y\}|
    \end{equation}
    where $\mathcal{Y}$ is the set of \gk{true class} labels. 
    The number of matching ground-truth and plural label assignments in cluster $j$ is:
    \begin{equation}
    m_j = |\{i \in \mathcal{C}_j : y_i = y_j^*\}|
    \end{equation}
    The clustering accuracy is then:
    \begin{equation}
    \text{ACC} = \frac{\sum_{j=1}^K m_j}{N}
    \end{equation}   
    where $K$ is the number of clusters and \gk{$N=\sum_j |\mathcal{C}_j|$ the total number of training samples.} This metric ranges from 0 to 1, with 1 indicating perfect clustering.

    \item \textbf{Normalized Mutual Information (NMI)}: This metric measures the information-theoretic agreement between true labels and predicted clusters. It quantifies how much information is shared between the two labelings, normalized by their individual entropies.
    The mutual information between true labels $\mathbf{y}$ and $K$ predicted clusters $\mathbf{c}$ is:
    \begin{equation}
    I(\mathbf{y}; \mathbf{c}) = \sum_{y \in \mathcal{Y}} \sum_{j=1}^K p(y, j) \log \frac{p(y, j)}{p_1(y) p_2(j)}
    \end{equation}  
    where
    \gk{$p(y, j) = |\{i : y_i = y,~  c_i = j\}|/N$ 
    is the joint probability, $p_1(y) = |\{i : y_i = y\}|/N$ is the marginal probability of true label $y$, and $p_2(j) = |\{i : c_i = j\}|/N$ is the marginal probability of cluster $j$.}
    
    The entropies are:
    \begin{align}
    H_1(\mathbf{y}) &= -\sum_{y \in \mathcal{Y}} p_1(y) \log p_1(y) \\
    H_2(\mathbf{c}) &= -\sum_{j=1}^K p_2(j) \log p_2(j)
    \end{align}  
    The normalized mutual information is:
    \begin{equation}
    \text{NMI} = \frac{I(\mathbf{y}; \mathbf{c})}{\sqrt{H_1(\mathbf{y}) H_2(\mathbf{c})}}
    \end{equation}    
    This metric ranges from 0 to 1, where 1 indicates perfect agreement and 0 indicates independence. NMI is symmetric and does not require label-to-cluster matching, making it robust to different numbers of clusters and classes.

    \item \textbf{Adjusted Rand Index (ARI).}
    this measures the pairwise similarity between two partitions while correcting for the agreement that would be expected by chance. ARI evaluates how many pairs of samples are clustered together in both labelings, compared to the expected number under a random assignment with the same cluster-size distribution.
    Let $n_{ij} = |\{k : y_k = i ,~ c_k = j\}|$ be the contingency table count of samples with true class $i$ and predicted cluster $j$. Let $a_i = \sum_j n_{ij}$ denote the size of true class $i$, and $b_j = \sum_i n_{ij}$ the size of predicted cluster $j$. The number of sample pairs that appear in the same class \emph{and} the same predicted cluster is:
    \begin{equation}
    A = \sum_{i,j} \binom{n_{ij}}{2}.
    \end{equation}  
    The expected number of such same agreements under random labeling (with fixed class sizes $\{a_i\}$ and cluster sizes $\{b_j\}$) is:
\gk{
    \begin{equation}
    B =
    \frac{
    \sum_i \binom{a_i}{2}
    \sum_j \binom{b_j}{2}
    }{
    \binom{N}{2}
    }.
    \end{equation}  
}
    The average number of same cluster pairs in the two partitions is:
    \begin{equation}
    C =
    \frac{1}{2}
    \left[
    \sum_i \binom{a_i}{2}
    +
    \sum_j \binom{b_j}{2}
    \right].
    \end{equation}
    The Adjusted Rand Index is then defined as:
    \begin{equation}
    \text{ARI} = \frac{A - B}{C - B}.
    \end{equation}
    ARI equals $1$ when the two partitions are identical, equals $0$ when the agreement matches the expected value under random labeling, and can be negative when the agreement is worse than random. ARI is symmetric and permutation invariant.

\squishend

\section{Autoencoder-Based Clustering}
\label{sec:method}

The approach studied herein is fully unsupervised, as outlined in
Section~2: an encoder is obtained as the front end of an autoencoder
trained for reconstruction only, and clustering is then performed by
$K$-means++ on the codewords it produces. The class labels play no
role at either stage and are used only to compute the metrics of
Section~3.

\subsection{FoldingNet Autoencoder}
\label{sec:foldingnet}

FoldingNet \cite{yang2018FoldingNet} consists of 
encoder $\phi$ and
two-stage folding decoder $\psi$.
%, jointly parameterized by weights $W$.
Given an input point cloud $x \in \mathbb{R}^{|x| \times 3}$ with $|x|$
points, the encoder produces a codeword $z = \phi(x) \in \mathbb{R}^{D}$
(we use $D=512$). The encoder maps each point through a three-layer MLP
to a 64-dimensional feature, aggregates local features by graph-based
max pooling ($64 \rightarrow 128 \rightarrow 1024$), max-pools over all
points to a global feature, and projects it to $\mathbb{R}^D$. Because
of the final max pooling the codeword is invariant to the order in
which the points are listed.
The decoder reconstructs a point cloud by two folding operations: a
fixed 2D grid $U \in \mathbb{R}^{M\times 2}$ ($M = 45^2 = 2025$ in our
experiments) is concatenated with the codeword and passed through an MLP
to produce an intermediate 3D point cloud, which is again concatenated
with the codeword and passed through a second MLP to produce the final
reconstruction $\hat{x} = \psi(z) \in \mathbb{R}^{M\times 3}$, with
$M=|\hat{x}|.$

The autoencoder is trained on the 80\% training split by minimizing the
bidirectional (squared) Chamfer distance between input and
reconstruction,
\begin{equation}\label{eq:chamfer}
d^{(2)}_{\rm c}(x, \hat x) = \frac{1}{|x|}\sum_{i=1}^{|x|}\min_{j}\|x_i - \hat{x}_j\|^2
+ \frac{1}{|\hat{x}|}
\sum_{j=1}^{|\hat{x}|}\min_{i}\|x_i - \hat{x}_j\|^2 ,
\end{equation}
i.e., the squared counterpart of the distance $d_{\rm c}$ of
Section~2. The reconstruction loss on the 20\% held-out split is
monitored during training; the checkpoint with the lowest held-out
loss is retained.

\subsection{K-Means++ on Codewords}
\label{sec:kmeanspp}

Once the autoencoder is trained, the decoder is discarded and every
shape $x\in X$ is mapped to its codeword $z=\phi(x)\in\mathbb{R}^D$.
Clustering is then performed in codeword space with the Euclidean
distance, i.e., the cluster-distortion loss $L_0$ of Section~2 is
minimized with $d_{\rm c}(x,\pi)$ replaced by $\|\phi(x)-\pi\|^2$ and
prototypes $\pi_1,\dots,\pi_K\in\mathbb{R}^D$:
\begin{equation}\label{eq:kmeans}
L(\pi) = \frac{1}{|X|}\sum_{x\in X}\min_{k=1,\dots,K}\|\phi(x)-\pi_k\|^2 .
\end{equation}
Lloyd's algorithm alternates assignment of each codeword to its nearest
prototype and recomputation of each prototype as the mean of the
codewords assigned to it. As noted in Section~2, $K$-means is prone to
cluster collapse and poor local minima when the prototypes are poorly
initialized; we therefore use $K$-means++ seeding \cite{Duda}, in
which each successive initial prototype is drawn from the codewords
with probability proportional to its squared distance to the nearest
prototype already chosen, so that initial prototypes are spread across
the data. The algorithm is run from 50 independent seedings (each for
up to 500 Lloyd iterations) and the solution with the lowest distortion
\eqref{eq:kmeans} is retained.

The number of clusters $K$ is not known a priori in the unsupervised
setting. We sweep $K$ over a range of values bracketing the number of
classes\\
($K\in\{2,4,\dots,20\}$ for MN10 and
$K\in\{10,20,30,35,40,45,50,60,70,80\}$ for MN40)\\
and examine the
clustering metrics as a function of $K$, which, together with the
``elbow'' heuristic or a BIC-type penalty \eqref{eq:BIC} on the
distortion, indicates a suitable range for $K$. \na{In the experiments
we apply the elbow heuristic of Section~2 to the metric curves: the
operating point is the $K$ at which the curves flatten---NMI and ARI
cease to improve and ACC thereafter gains only by splitting classes
that are already pure rather than by separating confused ones; it is
marked by a box in Tables~\ref{tab:kmeans_results_mn10} and
\ref{tab:kmeans_results_mn40}. Since NMI, ARI and ACC use the class
labels, this is a label-informed reading of the elbow. We also report,
without using the labels, the knee of the mean codeword distortion
\eqref{eq:kmeans} (the $K$ of maximum distance to the chord joining the
first and last points of the curve) and two BIC-type criteria: the
Gaussian-mixture BIC \eqref{eq:bic_gmm} and the additive penalty
\eqref{eq:BIC}.} Note that for
imbalanced datasets such as ModelNet10 and ModelNet40 the best $K$
need not equal the number of classes: small classes may be absorbed
into neighboring ones while large, geometrically diverse classes may
be split.

\section{Experimental Results}
\label{sec:results}

All experiments use the 80/20 re-splits of Section~\ref{sec:datasets}
(3,920 / 979 shapes for MN10 and 9,849 / 2,462 shapes for MN40),
2,048 input points per shape, and $D=512$. The same training protocol
and the same $K$-means++ settings are used for both datasets; only the
range of $K$ differs (Section~\ref{sec:kmeanspp}).

\subsection{Autoencoder Training}
\label{sec:ae_training}

We train one FoldingNet autoencoder of Section~\ref{sec:foldingnet} per dataset on its 80\% training split. The encoder takes as input 2,048 points uniformly sampled from each shape, and the decoder outputs a 2,025-point reconstruction, corresponding to the size of the 2D grid that is ``folded'' into the target shape. The minimizing objective of the autoencoder is the averaged bidirectional Chamfer distance \eqref{eq:chamfer} between the source and target shape. Each autoencoder was trained for 100 epochs with Adam \cite{kingma2014adam} (learning rate $10^{-3}$, halved every 25 epochs, batch size 32).

\begin{figure}[!tb]
    \centering
    \begin{subfigure}[b]{\columnwidth}
        \centering
        \includegraphics[width=\linewidth]{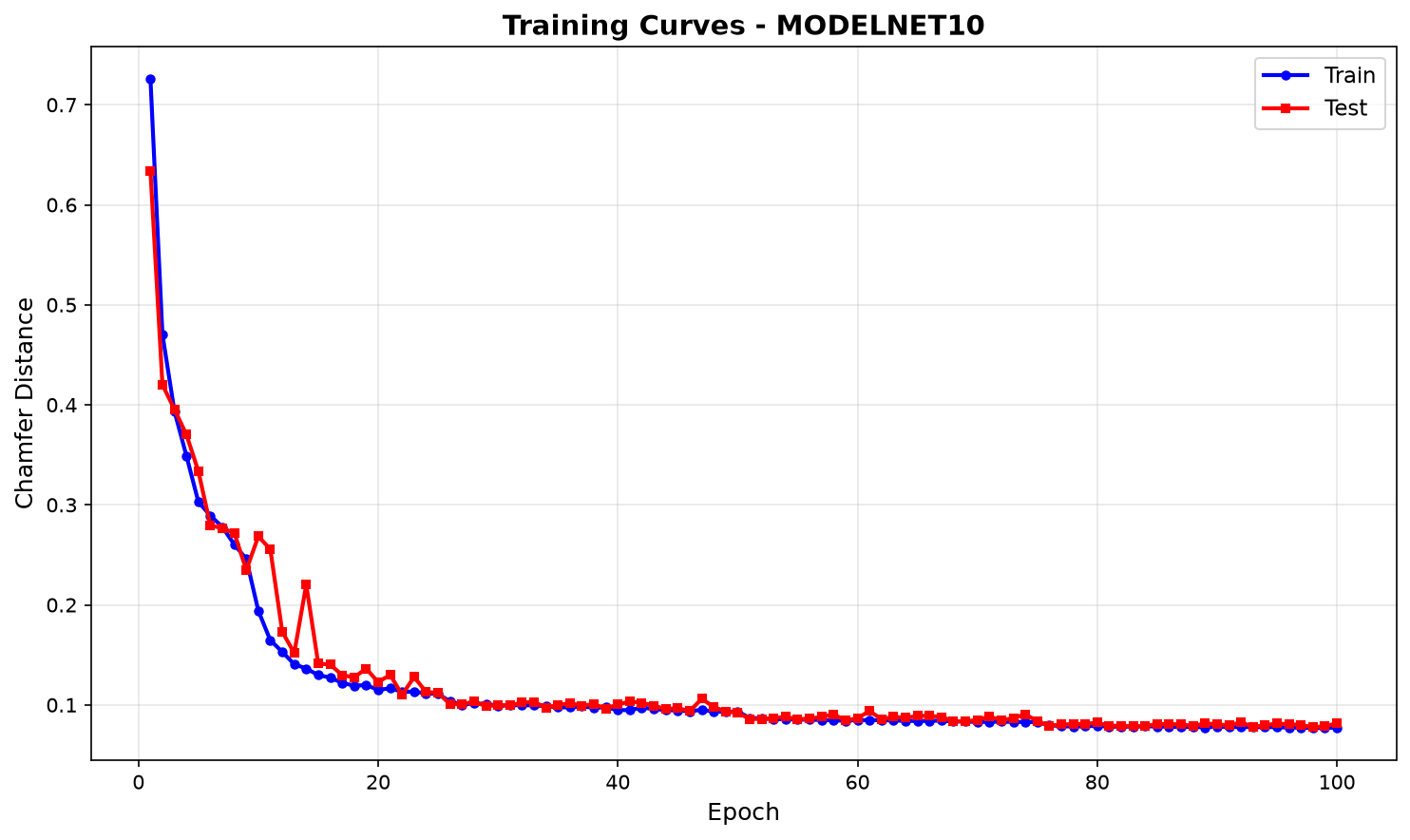}
        \caption{Chamfer distance vs.\ epoch (training and held-out splits).}
    \end{subfigure}\\[4pt]
    \begin{subfigure}[b]{\columnwidth}
        \centering
        \includegraphics[width=\linewidth]{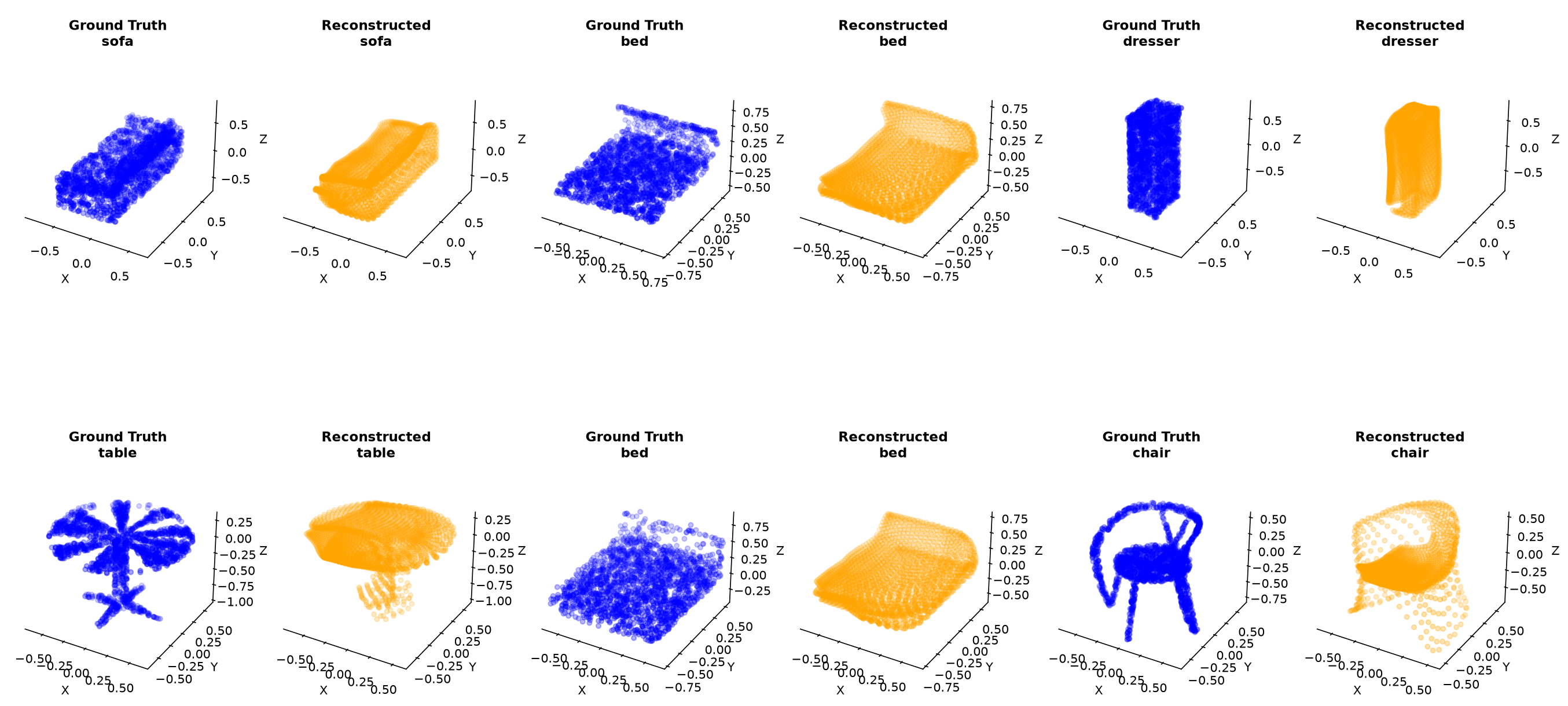}
        \caption{Example shapes (blue) and their FoldingNet reconstructions (orange).}
    \end{subfigure}
    \caption{Training of the FoldingNet autoencoder on ModelNet10.}
    \label{fig:ae_training_mn10}
\end{figure}

\paragraph{ModelNet10.}
Figure~\ref{fig:ae_training_mn10} shows the training and held-out Chamfer distance over epochs and some example reconstructions. \na{The held-out loss tracks the training loss closely after about 30 epochs, indicating no overfitting of the encoder. The final Chamfer distance (reconstruction loss before applying KMeans++) was \MNTenAETrainLoss\ on the training split and \MNTenAETestLoss\ on the held-out split; the checkpoint with the lowest held-out loss, reached at epoch \MNTenAEBestEpoch\ of \MNTenAEEpochs, is used in the sequel.} The reconstructions capture the global shape of each object (the overall extent of a bed or sofa, the top of a table, the back and seat of a chair) but smooth away thin structures such as table legs and chair spindles, which is typical of folding-based decoders with a single codeword.
%but attempting to cluster based on its
%encoder will result in high cluster distortion loss.

\begin{figure}[!tb]
    \centering
    \begin{subfigure}[b]{\columnwidth}
        \centering
        \includegraphics[width=\linewidth]{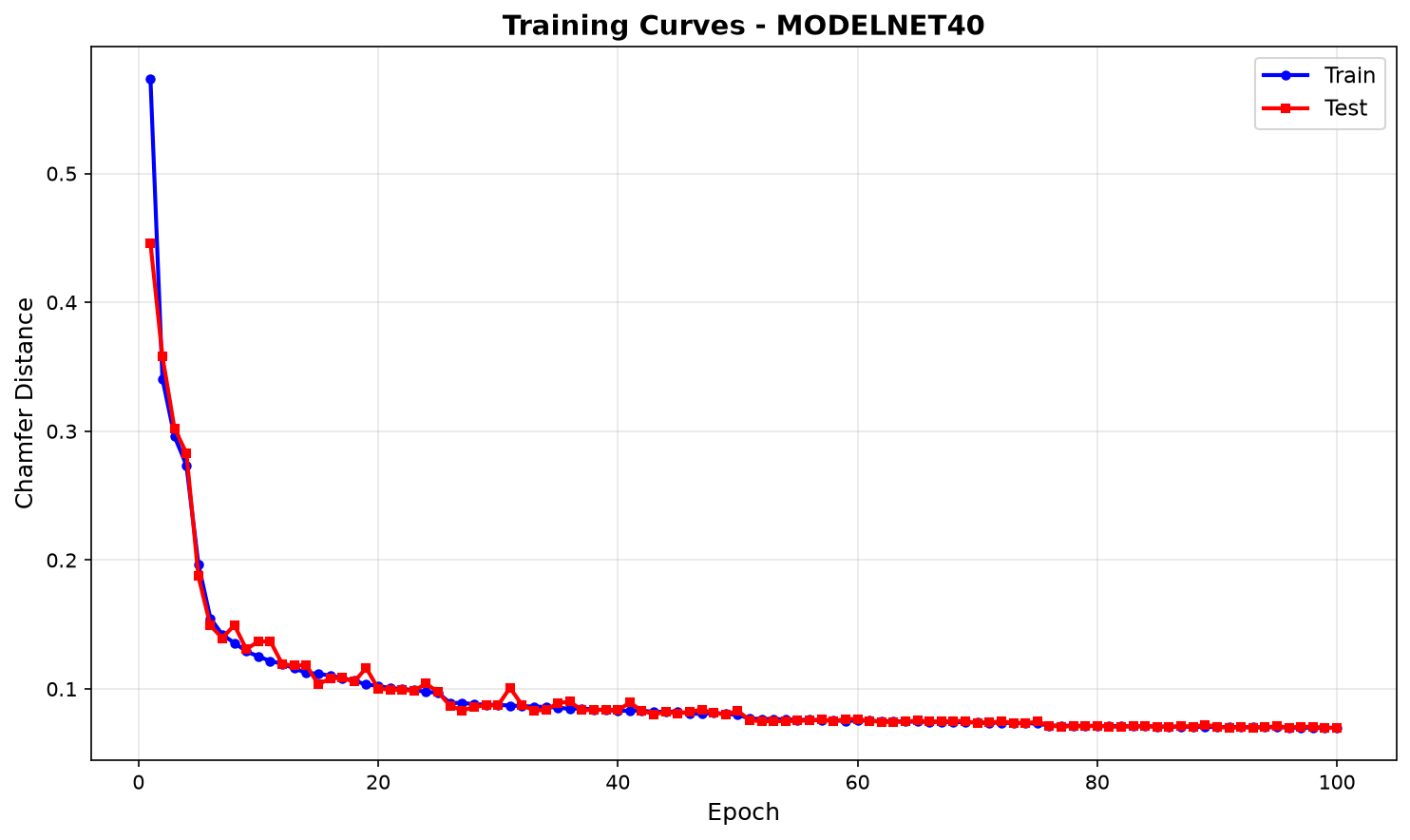}
        \caption{Chamfer distance vs.\ epoch (training and held-out splits).}
    \end{subfigure}\\
    \begin{subfigure}[b]{\columnwidth}
        \centering
        \includegraphics[width=\linewidth]{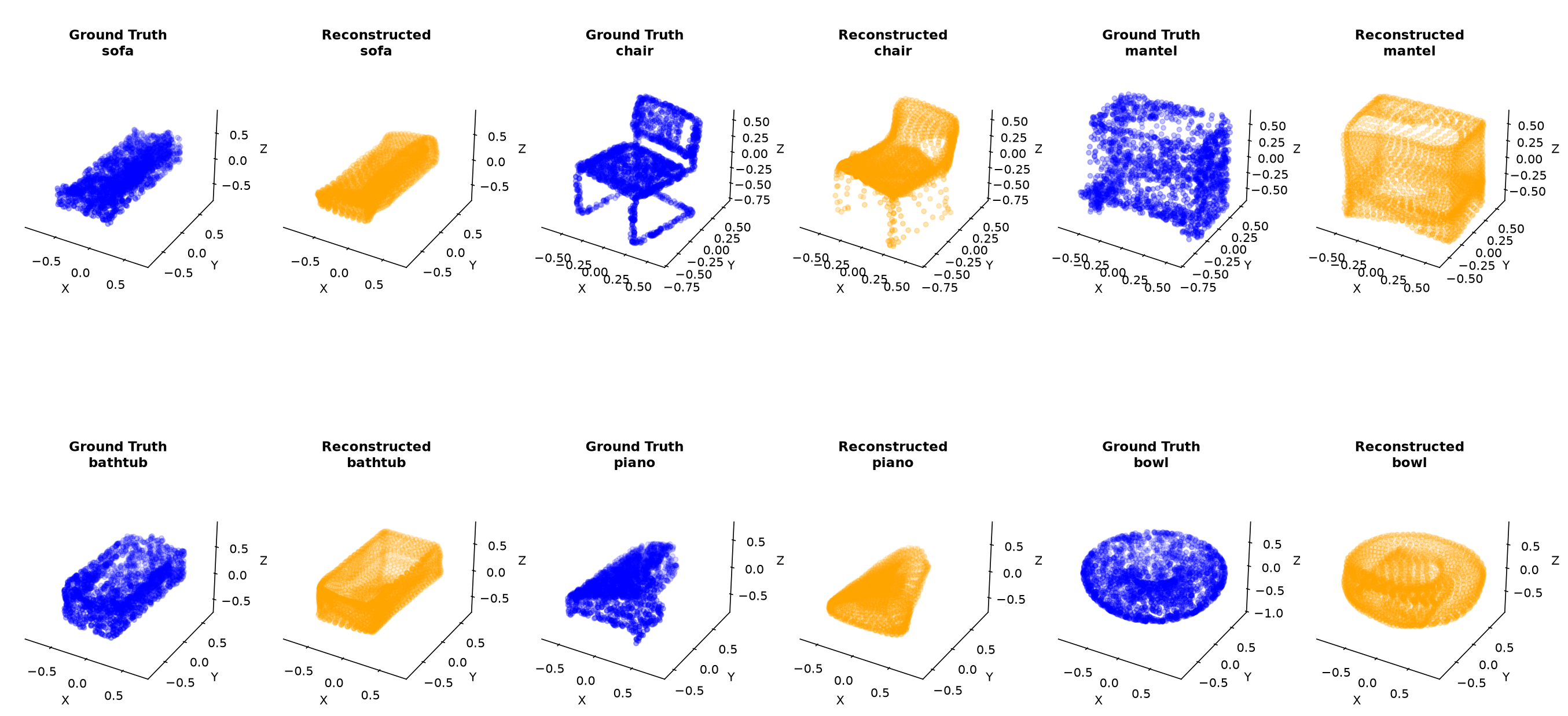}
        \caption{Example shapes (blue) and their FoldingNet reconstructions (orange).}
    \end{subfigure}
    \caption{\na{Training of the FoldingNet autoencoder on ModelNet40.}}
    \label{fig:ae_training_mn40}
\end{figure}

\paragraph{ModelNet40.}
Figure~\ref{fig:ae_training_mn40} shows the corresponding curves and
reconstructions for MN40. \na{The final held-out Chamfer distance was
\MNFortyAETestLoss\ (training \MNFortyAETrainLoss), with the lowest
held-out loss reached at the last epoch (\MNFortyAEBestEpoch\ of
\MNFortyAEEpochs). With 2.5 times more training shapes, the two curves
are smoother than on MN10 and coincide after about 30 epochs, so there
is again no overfitting of the encoder; the final loss is in fact
slightly lower than on MN10, since the larger training set outweighs
the greater diversity of the 40 categories. The reconstructions show
the same behavior as on MN10: the coarse volume of a sofa, bathtub,
mantel or bowl is reproduced faithfully, whereas the legs of a chair
or the lid and legs of a piano are smoothed into a solid block, and
categories with thin or highly articulated structures (plant, lamp,
person, guitar, airplane) are reconstructed as their convex envelope.}

\begin{figure*}[!t]
    \centering
    \includegraphics[width=\textwidth]{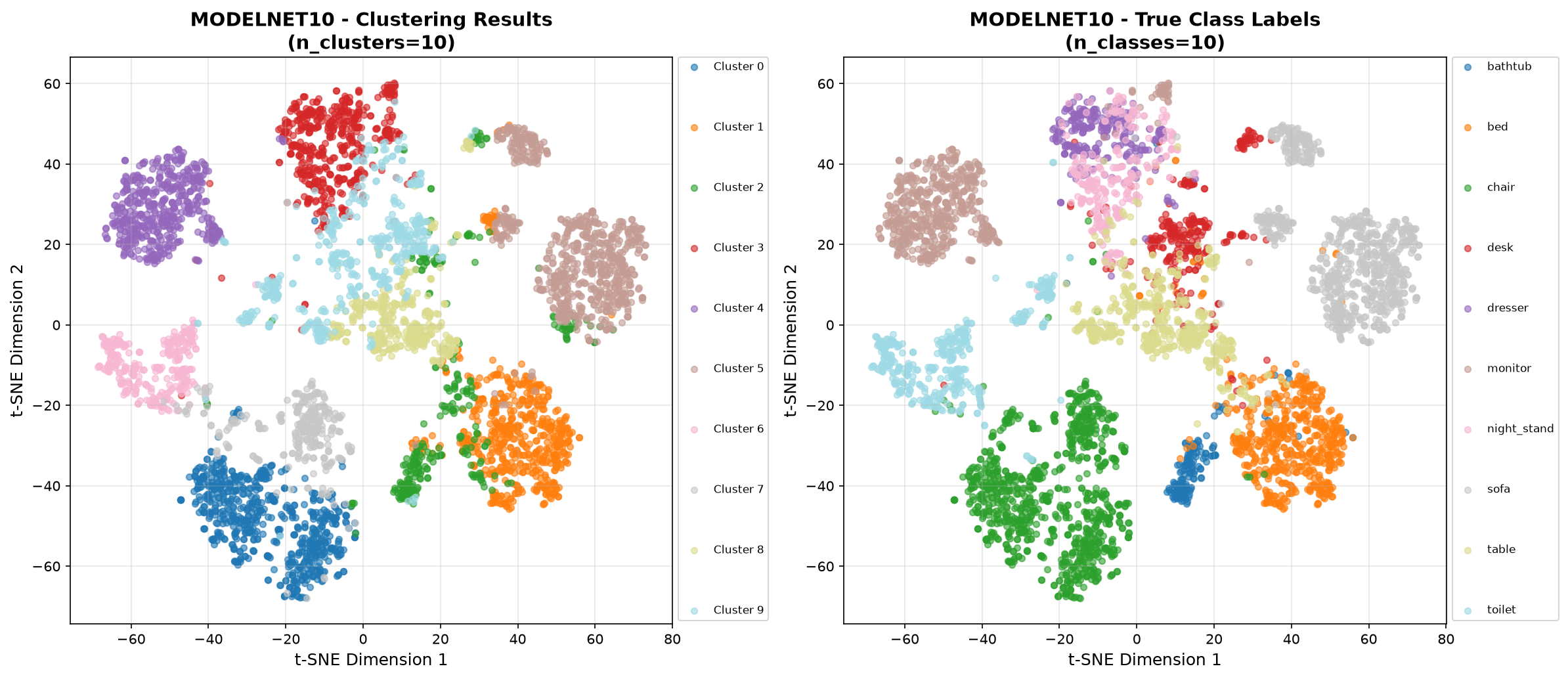}
    \caption{\na{t-SNE projection of FoldingNet codewords for ModelNet10, colored by KMeans++ cluster ($K=10$, left) and by true class (right). Monitor, toilet, bathtub and the sub-groups of chair, sofa and bed form compact, well separated groups, whereas desk/table and night stand/dresser overlap in the center of the projection.}}
    \label{fig:kmeans_tsne_mn10}
\end{figure*}

\subsection{K-Means++ Clustering of Codewords: ModelNet10}
\label{sec:kmeans_mn10}

\begin{figure}[!tb]
    \centering
    \includegraphics[width=\columnwidth]{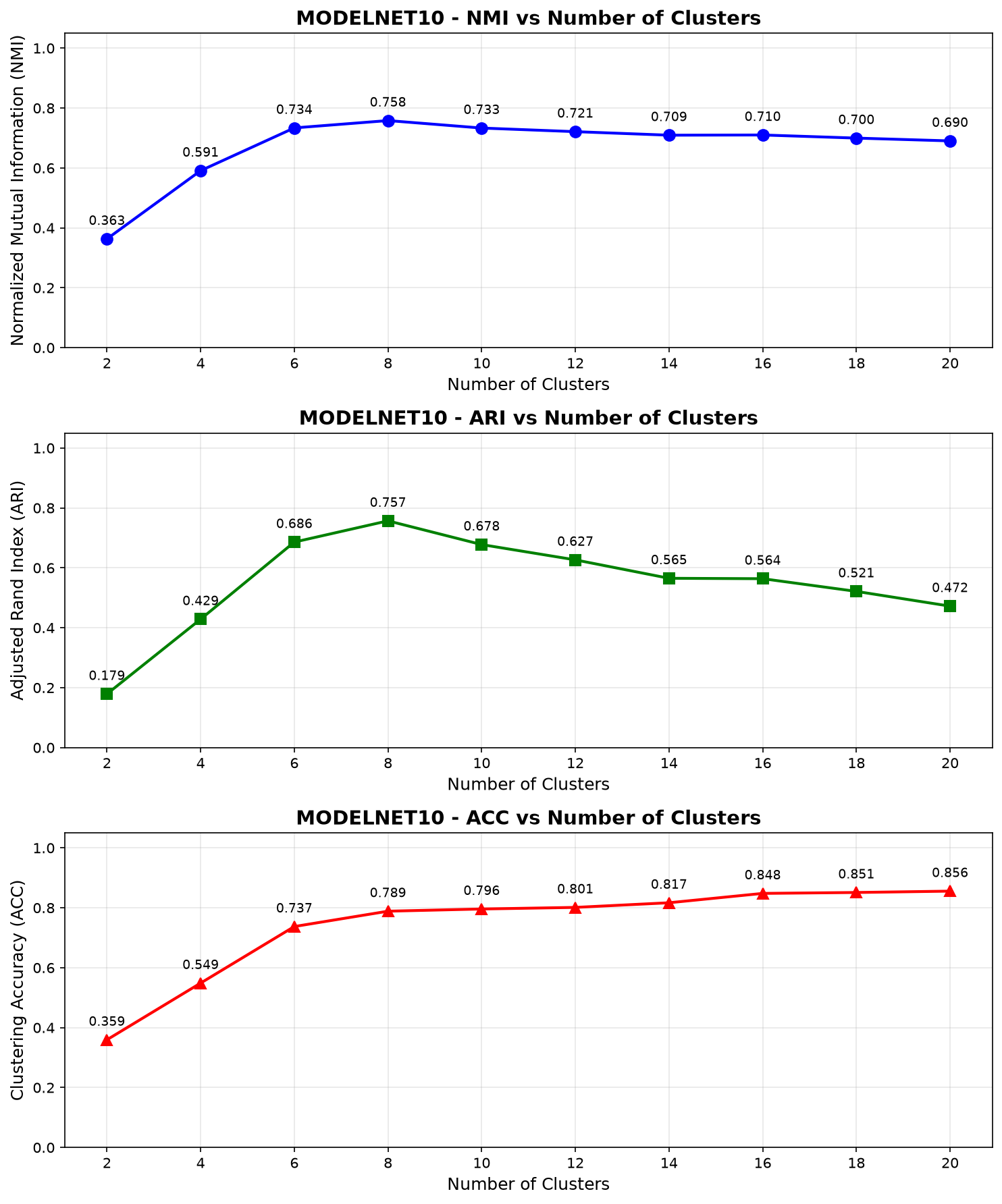}
    \caption{\na{NMI, ARI and ACC of KMeans++ on FoldingNet codewords vs.\ the number of clusters $K$ (ModelNet10). NMI and ARI peak at $K=8$; ACC keeps increasing up to $K=20$.}}
    \label{fig:kmeans_vs_K_mn10}
\end{figure}

Table~\ref{tab:kmeans_results_mn10} and Figure~\ref{fig:kmeans_vs_K_mn10} show clustering results for K-Means++ (50 restarts, up to 500 iterations each) on the encoder codewords of all 4,899 shapes, for $K$ ranging from 2 to 20.

\begin{table}[ht]
\centering
\caption{\na{Autoencoding/K-Means++ Clustering Results on ModelNet10 (all 4,899 shapes). Best value per column in bold; the boxed $K$ is the elbow (operating point).}}
\label{tab:kmeans_results_mn10}
\na{
\renewcommand\elbowK{8}%
\begin{tabular}{|E|c|c|c|}
\hline
\multicolumn{1}{|c|}{\textbf{\# Clusters} ($K$)} & \textbf{NMI} & \textbf{ARI} & \textbf{ACC} \\
\hline
2 & 0.363 & 0.179 & 0.359 \\
4 & 0.591 & 0.429 & 0.549 \\
6 & 0.734 & 0.686 & 0.737 \\
8 & \textbf{0.758} & \textbf{0.757} & 0.789 \\
10 & 0.733 & 0.678 & 0.796 \\
12 & 0.721 & 0.627 & 0.801 \\
14 & 0.709 & 0.565 & 0.817 \\
16 & 0.710 & 0.564 & 0.848 \\
18 & 0.700 & 0.521 & 0.851 \\
20 & 0.690 & 0.472 & \textbf{0.856} \\
 \hline
\end{tabular}}
\end{table}

\begin{figure}[!tb]
    \centering
    \includegraphics[width=\columnwidth]{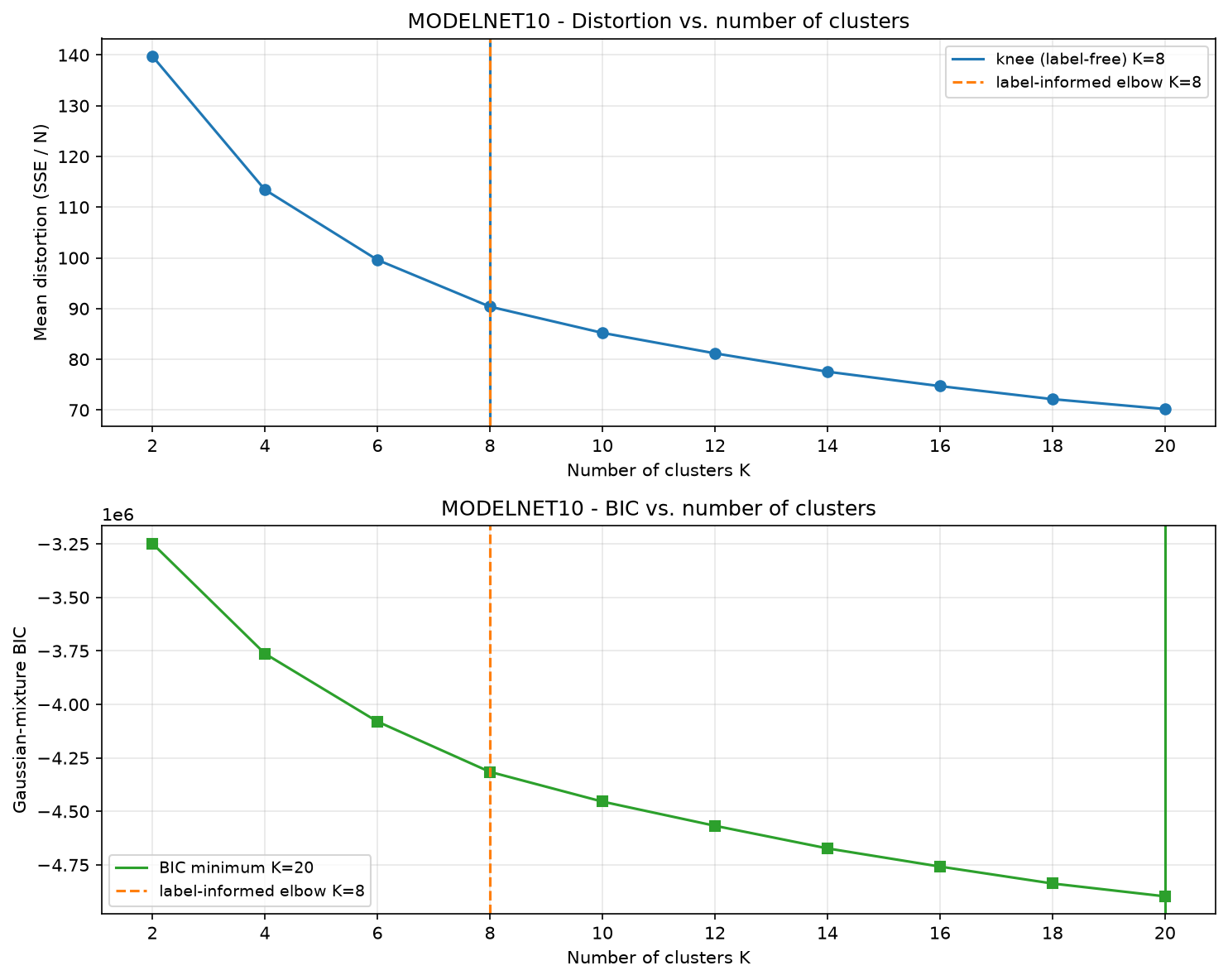}
    \caption{\na{Mean codeword distortion (top) and Gaussian-mixture BIC \eqref{eq:bic_gmm} (bottom) vs.\ $K$ on ModelNet10. The solid line marks the label-free knee $K=\MNTenKneeK$, which coincides with the boxed elbow; \eqref{eq:bic_gmm} has no interior minimum and selects $K=\MNTenBICK$.}}
    \label{fig:kmeans_distortion_mn10}
\end{figure}

\na{
\paragraph{Effect of the number of clusters.}
All three metrics increase steeply from $K=2$ to $K=8$, as the coarse
partitions for small $K$ are forced to merge dissimilar classes. NMI and
ARI peak at $K=8$ (NMI 0.758, ARI 0.757, ACC 0.789 in
Table~\ref{tab:kmeans_results_mn10}), i.e., slightly below the number of
classes, and the three metrics then behave differently for larger $K$.
ARI decreases monotonically beyond $K=8$ (to 0.472 at $K=20$) because
ARI penalizes splitting a true class across several clusters: the large
and geometrically diverse classes (chair, sofa, bed) are the first to be
over-segmented as $K$ grows. NMI, by contrast, is nearly flat for
$K\ge 8$ (between 0.690 and 0.758), since splitting a class into pure
sub-clusters costs little mutual information. ACC, which only credits
the plurality class of each cluster, is not penalized by such splits at
all and therefore keeps increasing with $K$, from 0.789 at $K=8$ to
\MNTenBestACC\ at $K=\MNTenBestK$ (NMI \MNTenBestNMI, ARI
\MNTenBestARI), the end of the sweep; the gain, however, comes from
additional clusters being carved out of already-pure classes (at
$K=20$, chair occupies three clusters of 90--100\% purity, sofa four,
bed two, table three and monitor two) rather than from separating the
confused ones. Taken together, the curves exhibit a clear elbow at
$K=8$, which we take as the operating point of the autoencoder/KMeans++
pipeline on ModelNet10; this is consistent with the heuristics
discussed in Section~2 and with the class imbalance of the dataset: the
two ``missing'' clusters correspond to the small classes night stand
and desk (286 shapes each), which are merged with their larger,
geometrically similar neighbors dresser and table.

\paragraph{Label-free choice of $K$.}
Figure~\ref{fig:kmeans_distortion_mn10} shows the mean distortion
\eqref{eq:kmeans} and the Gaussian-mixture BIC \eqref{eq:bic_gmm}
against $K$. The knee of the distortion curve is at $K=\MNTenKneeK$,
which coincides with the boxed, label-informed elbow. Both BIC-type
criteria select $K=\MNTenBICK$, the end of the sweep: mean distortion
keeps falling, so neither \eqref{eq:bic_gmm} nor the additive penalty
\eqref{eq:BIC} has an interior minimum on these 512-dimensional
codewords.

\paragraph{Which classes are separated.}
Figure~\ref{fig:kmeans_tsne_mn10} shows a t-SNE projection of the codewords
for $K=10$, colored by cluster assignment (left) and by true class
(right). Classes with a distinctive global geometry, such as monitor,
toilet, chair, bed and bathtub, form compact, well separated groups that
are recovered almost exactly by KMeans++: at $K=20$, the plurality
clusters of monitor (two clusters, both 100\% pure, together containing
92\% of the class), chair (three clusters of 90--100\%), toilet (99\%),
bed (93\% and 83\%), sofa (four clusters of 89--98\%) and bathtub
(97\%) are all dominated by a single class. The errors concentrate on
pairs of classes with similar overall shape: desk and table (a flat top
with supports) are merged into a common region of codeword space (7\%
of desks fall into the main table cluster and 13\% of tables into the
main desk cluster, which is only 61\% pure), as are night stand and
dresser (box-like shapes differing mainly in aspect ratio; night
stand's two plurality clusters are only 43--54\% pure and a quarter of
its shapes fall into the main dresser cluster), while bathtub leaks
into the bed clusters (23\% of the class; elongated, low, hollowed
shapes). Because the codeword is learned only to minimize the
reconstruction error \eqref{eq:chamfer}, it encodes the coarse volume
of the object faithfully while thin structures, which often carry the
class-discriminative information (e.g., the drawers of a dresser or the
legs of a table), are smoothed away in the reconstructions of
Figure~\ref{fig:ae_training_mn10}; this limits how well these class
pairs can be separated by any clustering of the codewords. Such
confusion is also consistent with the elbow at $K=8$: the two merged
pairs account for the two ``missing'' clusters relative to the ten
classes.
}

\begin{figure*}[!t]
    \centering
    \includegraphics[width=\textwidth]{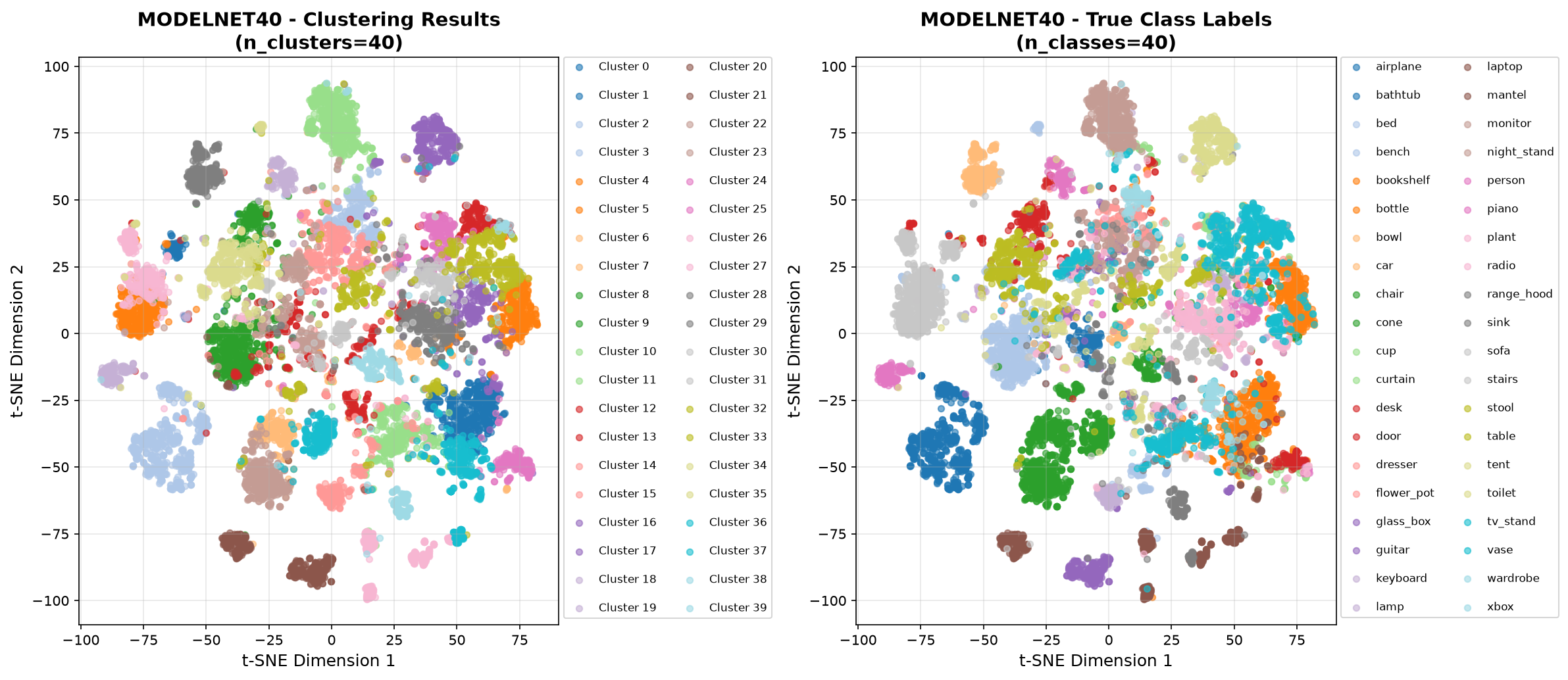}
    \caption{\na{t-SNE projection of FoldingNet codewords for ModelNet40, colored by KMeans++ cluster ($K=40$, left) and by true class (right). Airplane, car, guitar, laptop, chair, toilet and monitor form compact, well separated groups, whereas the box-like furniture classes (night stand, dresser, tv stand, glass box, wardrobe, xbox) and the container-like classes (vase, cup, flower pot, plant) overlap in the center of the projection.}}
    \label{fig:kmeans_tsne_mn40}
\end{figure*}

\subsection{K-Means++ Clustering of Codewords: ModelNet40}
\label{sec:kmeans_mn40}

\begin{figure}[!tb]
    \centering
    \includegraphics[width=\columnwidth]{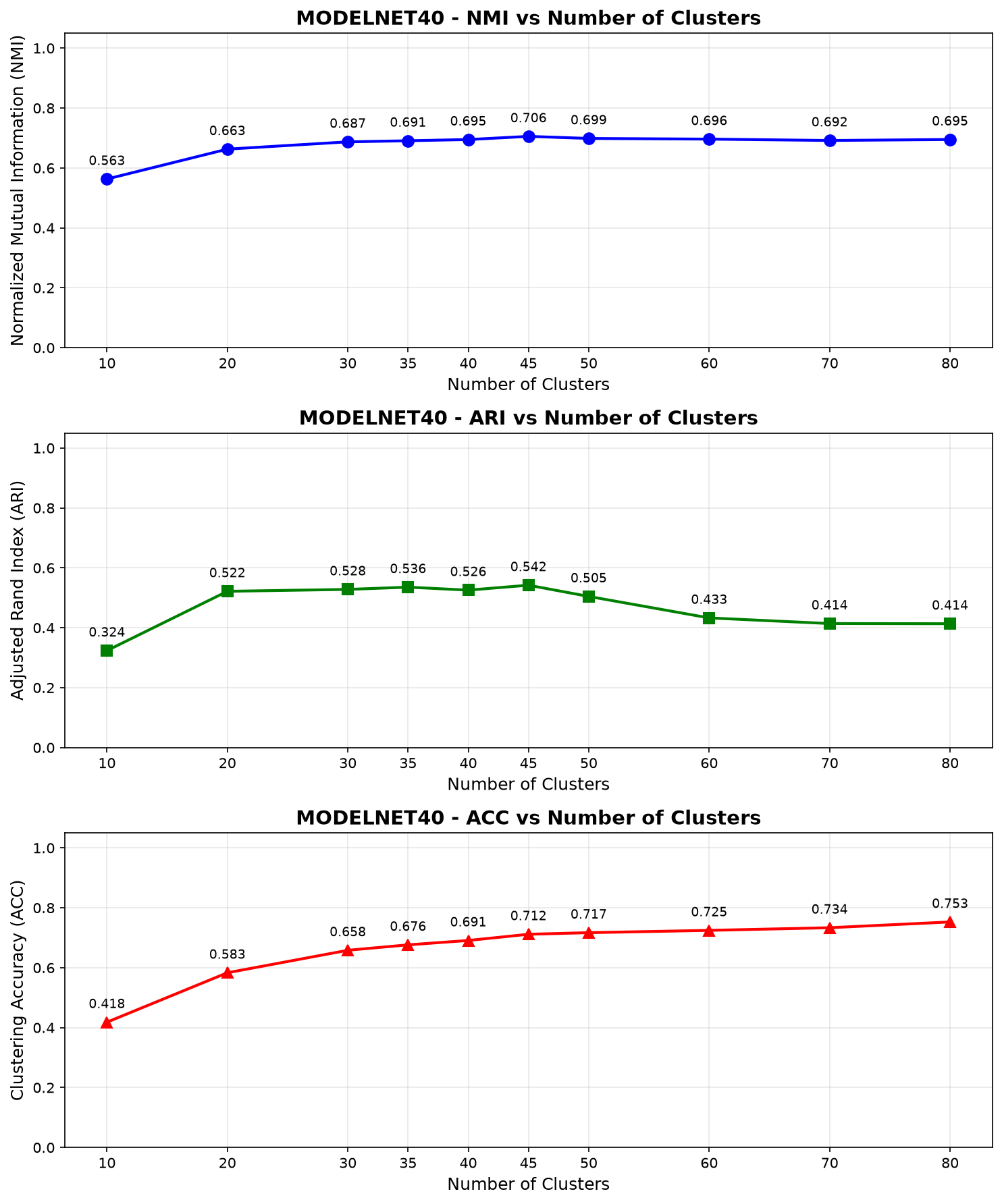}
    \caption{\na{NMI, ARI and ACC of KMeans++ on FoldingNet codewords vs.\ the number of clusters $K$ (ModelNet40). NMI and ARI peak at $K=45$; the boxed operating point remains $K=30$, past which ACC improves only slowly up to $K=80$.}}
    \label{fig:kmeans_vs_K_mn40}
\end{figure}

\na{Figure~\ref{fig:kmeans_tsne_mn40} shows the t-SNE projection of the
MN40 codewords for $K=40$. Table~\ref{tab:kmeans_results_mn40} and
Figure~\ref{fig:kmeans_vs_K_mn40} show the corresponding clustering
results: KMeans++ (50 restarts, up to 500 iterations each) on the
codewords of all 12,311 shapes, for $K$ ranging from 10 to 80.}

\begin{table}[ht]
\centering
\caption{\na{Autoencoding/K-Means++ Clustering Results on ModelNet40 (all 12,311 shapes). Best value per column in bold; the boxed $K$ is the elbow (operating point).}}
\label{tab:kmeans_results_mn40}
\na{
\renewcommand\elbowK{30}%
\begin{tabular}{|E|c|c|c|}
\hline
\multicolumn{1}{|c|}{\textbf{\# Clusters} ($K$)} & \textbf{NMI} & \textbf{ARI} & \textbf{ACC} \\
\hline
10 & 0.563 & 0.324 & 0.418 \\
20 & 0.663 & 0.522 & 0.583 \\
30 & 0.687 & 0.528 & 0.658 \\
35 & 0.691 & 0.536 & 0.676 \\
40 & 0.695 & 0.526 & 0.691 \\
45 & \textbf{0.706} & \textbf{0.542} & 0.712 \\
50 & 0.699 & 0.505 & 0.717 \\
60 & 0.696 & 0.433 & 0.725 \\
70 & 0.692 & 0.414 & 0.734 \\
80 & 0.695 & 0.414 & \textbf{0.753} \\
 \hline
\end{tabular}}
\end{table}

\begin{figure}[!tb]
    \centering
    \includegraphics[width=\columnwidth]{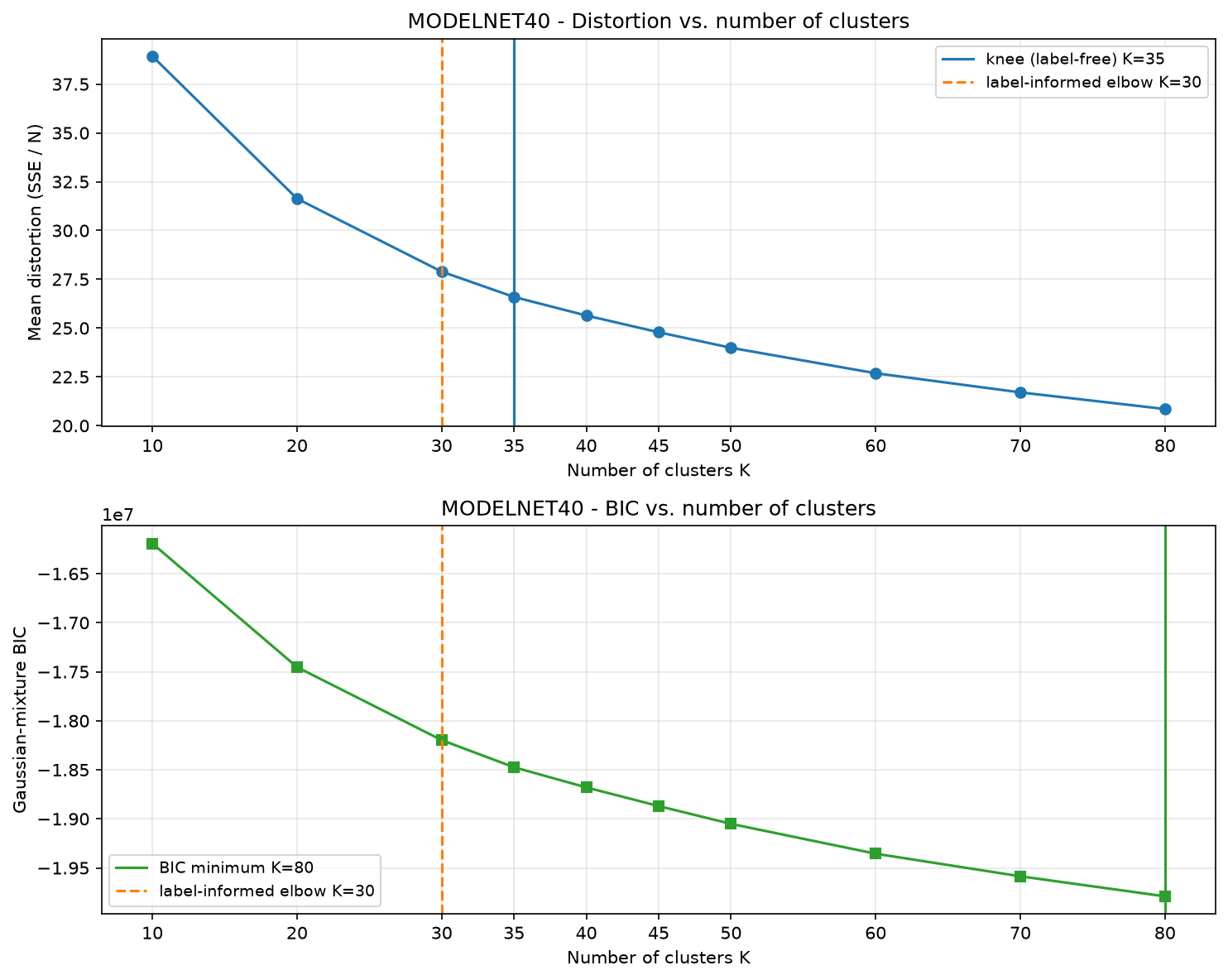}
    \caption{\na{Mean codeword distortion (top) and Gaussian-mixture BIC \eqref{eq:bic_gmm} (bottom) vs.\ $K$ on ModelNet40. The solid line marks the label-free knee $K=\MNFortyKneeK$; the dashed line is the boxed elbow $K=30$. \eqref{eq:bic_gmm} has no interior minimum and selects $K=\MNFortyBICK$.}}
    \label{fig:kmeans_distortion_mn40}
\end{figure}

\na{
\paragraph{Effect of the number of clusters.}
All three metrics rise steeply from $K=10$ to $K=30$: at $K=10$ the 40
classes are necessarily merged into ten coarse families (NMI 0.563, ARI
0.324, ACC 0.418). At the boxed operating point $K=30$ (NMI 0.687, ARI
0.528, ACC 0.658), ten clusters below the number of classes, the steep
rise is over. NMI and ARI continue to a modest peak at $K=45$ (NMI
0.706, ARI 0.542) and then ARI falls (to 0.414 at $K=80$), because the
large, geometrically diverse classes are over-segmented: at $K=80$,
chair occupies six clusters, sofa, mantel, vase and piano four each,
and airplane, bed, bookshelf, monitor and table three each. NMI is
flat for $K\ge 30$ (between 0.687 and 0.706), since splitting a class
into pure sub-clusters costs little mutual information. ACC, which
only credits the plurality class of each cluster, continues to
increase with $K$ but with diminishing returns: it gains 0.10 (from
0.658 to 0.753) when $K$ is increased from 30 to 80, since the
additional clusters are mostly carved out of already-pure classes
rather than separating the confused ones. The best measured ACC is
\MNFortyBestACC\ (NMI \MNFortyBestNMI, ARI \MNFortyBestARI) at
$K=\MNFortyBestK$, the end of the sweep. We keep $K=30$ as the
operating point on ModelNet40: it is where the metric curves flatten,
and the later NMI/ARI peak at $K=45$ is a small gain on an already
flat plateau.

\paragraph{Label-free choice of $K$.}
Figure~\ref{fig:kmeans_distortion_mn40} shows the mean distortion
\eqref{eq:kmeans} and the Gaussian-mixture BIC \eqref{eq:bic_gmm}
against $K$. The knee of the distortion curve is at $K=\MNFortyKneeK$,
five clusters above the boxed elbow. The additive penalty
\eqref{eq:BIC} selects $K=\MNFortyPenalizedK$, which coincides with the
boxed operating point. The Gaussian-mixture BIC \eqref{eq:bic_gmm}
again has no interior minimum and selects $K=\MNFortyBICK$, the end of
the sweep. Relative to
MN10, the metric curves are flatter, so the label-informed peak
($K=45$), the distortion knee ($K=35$) and the boxed elbow ($K=30$) no
longer coincide.

\paragraph{Which classes are separated.}
Figure~\ref{fig:kmeans_tsne_mn40} shows the t-SNE projection of the
codewords for $K=40$. As on MN10, the classes with a distinctive global
geometry are recovered almost exactly: at $K=80$ the plurality clusters
of airplane (three clusters, each 100\% pure), car (98\%, containing
99\% of the class), guitar (100\%, containing 91\% of the class),
laptop (98\%), range hood (97\%), chair (six clusters, four of them
94--100\% pure), monitor (99\%, 83\% and 94\%), toilet (100\% and
69\%), mantel (four clusters, three of them 95--99\%), sofa (four
clusters of 82--98\%), piano (100\% and 86\%), bookshelf (95\% and
81\%), bed (93\% and 88\%), keyboard (86\%), bench (84\% and 95\%),
bathtub (82\%), stairs (81\% and 96\%), bowl (80\%), bottle (83\% and
73\%) and table (96\%, 85\% and 70\%) are all dominated by a single
class. Such confusion concentrates on families of classes with similar
coarse volume. The box-like furniture classes are the largest such
family: dresser's plurality cluster is only 57\% pure and is shared
with night stand (which loses 22\% of its shapes to it); night stand
itself wins three clusters of only 24--42\% purity; xbox and wardrobe
each win a mixed cluster of 45--49\% purity, and glass box, tv stand
and radio share clusters in which no class exceeds 49\%. The
container-like classes form the second family: cup wins a cluster that
is only 41\% cup and 35\% vase (and contains 66\% of the cups), flower
pot is absorbed by plant and vase, and lamp is scattered over clusters
shared with plant, vase and stool. 
The flat, upright panels of curtain
and door form the third family (their plurality clusters are 48--50\%
pure and each contains the other class).
Less confused are:
desk with table, stool with night stand \& lamp, tent with cone.
Sink is scattered over many clusters. 
As a result, three
classes (flower pot, radio and stool) never obtain a cluster in which
they form the plurality, even at $K=80$, and most clusters won by
increasing $K$ beyond 30 are sub-clusters of chair, sofa, mantel,
airplane, bed or bookshelf. These merged families are exactly the
fine-grained categories that distinguish MN40 from MN10, and since the
codeword is learned only to minimize the Chamfer distance
\eqref{eq:chamfer}, the class-discriminative thin structures (drawers,
handles, the leaves of a plant) are smoothed away in reconstruction,
which limits how well these families can be separated by any clustering
of the codewords. Their number accounts for the gap between the elbow
at $K=30$ and the 40 classes.
}

\subsection{Comparison Across Datasets}
\label{sec:comparison}

\na{
The two datasets are processed with an identical pipeline, so the
differences in Tables~\ref{tab:kmeans_results_mn10} and
\ref{tab:kmeans_results_mn40} reflect the datasets rather than the
method. On MN10 the autoencoder/KMeans++ pipeline attains its best NMI
and ARI (0.76 and 0.76, with ACC 0.79) at the elbow $K=8$, two clusters
below the number of classes, and the errors are concentrated on two
pairs of confusable classes; the highest ACC (\MNTenBestACC, with NMI
\MNTenBestNMI\ and ARI \MNTenBestARI) is reached at $K=\MNTenBestK$ by
over-segmenting chair, sofa and bed. On MN40 the elbow is at $K=30$
(NMI 0.69, ARI 0.53, ACC 0.66), ten clusters below the number of
classes, and the highest ACC (\MNFortyBestACC, with NMI \MNFortyBestNMI\
and ARI \MNFortyBestARI) is likewise only reached at $K=\MNFortyBestK$
by over-segmenting the large classes. All three metrics are lower on
MN40 at the respective elbows (NMI 0.69 vs.\ 0.76, ARI 0.53 vs.\ 0.76,
ACC 0.66 vs.\ 0.79), the drop being smallest for NMI and largest for
ARI, and the gap between the elbow and the number of classes is larger
(10 vs.\ 2 clusters). The label-free distortion knee agrees with the
boxed elbow on MN10 ($K=\MNTenKneeK$) and is off by five clusters on
MN40 ($K=\MNFortyKneeK$ vs.\ $K=30$); the Gaussian-mixture BIC
\eqref{eq:bic_gmm} has no interior minimum on either 512-dimensional
sweep. This is because MN40
contains many more confusable families and more small classes (flower
pot, radio, stool, and to a lesser extent cup, xbox, wardrobe, tent,
lamp, curtain and door) that are absorbed by larger, geometrically
similar ones. On the other hand the ACC curves end at similar values
($0.86$ and $0.75$ at $K=2\times$ the number of classes), since
over-segmentation is not penalized by ACC and the large classes of both
datasets split into pure sub-clusters. In both cases the limiting
factor is the same: a codeword trained for reconstruction alone encodes
coarse volume faithfully but discards the thin, class-discriminative
structures, so geometrically similar categories are not separable in
codeword space regardless of $K$.
This effect is most pronounced on
MN40 where the confusable families are most numerous.
}

\section{Summary and On-Going Work}
\gk{
This paper concerns deep clustering of three-dimensional shapes
specified by point clouds. We studied the fully unsupervised
pipeline in which a feature map (encoder) is obtained through an
autoencoding mechanism, which avoids the cluster collapse to which
deep clustering is prone, and $K$-means clustering is then applied
to the encodings: a FoldingNet encoder is trained as part of an
autoencoder for reconstruction only and $K$-means++ is applied to
its codewords. Experimental results are provided for the ModelNet10
and ModelNet40 point cloud datasets.
\na{On ModelNet10 the clustering metrics exhibit a clear elbow at $K=8$
clusters (NMI 0.76, ARI 0.76, ACC 0.79), slightly below the number of
classes, with classes of distinctive global geometry (monitor, toilet,
chair, bed, bathtub) recovered almost exactly and geometrically similar
pairs (desk/table, night stand/dresser) merged in codeword space;
ACC continues to rise to \MNTenBestACC\ at $K=\MNTenBestK$ as the large
classes are split into pure sub-clusters. On ModelNet40 the elbow is at
$K=30$ (NMI 0.69, ARI 0.53, ACC 0.66), ten clusters below the 40
classes, with ACC rising only to \MNFortyBestACC\ at $K=\MNFortyBestK$;
all three metrics are lower than on ModelNet10 because the more
numerous fine-grained families (night stand/dresser/xbox/wardrobe,
glass box/tv stand/radio, vase/cup/flower pot/plant, curtain/door)
are merged and three small classes (flower pot, radio, stool) never
obtain their own cluster. The label-free knee of the codeword
distortion \eqref{eq:kmeans} coincides with the boxed elbow on
ModelNet10 ($K=\MNTenKneeK$) and is off by five clusters on ModelNet40
($K=\MNFortyKneeK$); the Gaussian-mixture BIC \eqref{eq:bic_gmm} has
no interior minimum on either sweep.}
In both cases the limiting factor is that a codeword trained for
reconstruction alone encodes coarse volume faithfully but smooths
away the thin structures that distinguish geometrically similar
categories. Training the autoencoder is also the dominant
computational cost of the pipeline, which matters for large-scale
datasets.
Future work within this pipeline includes studying the effect of the
codeword dimension $D$, the number of input points and early stopping
of the autoencoder on the clustering metrics, and evaluating the
pipeline on larger point-cloud datasets.
}

In on-going work, we are considering joint semisupervised learning
of encoder and $K$ prototypes  with pairwise contrastive loss penalties
at the cluster level. Such penalties may improve performance on
datasets with classes of high size variance and 
significant confusion among them.
We are also considering mulitmodal encoders including a 
point-cloud modality.

\section*{Acknowledgements}
This research was supported by NSF CMMI grant 2415752.


\bibliographystyle{plain}
%\bibliography{bibfiles/shape_clustering,bibfiles/adversarial2,bibfiles/manufacturing,bibfiles/ml}

\end{document}